\RequirePackage{fix-cm}
\documentclass[twocolumn]{svjour3}          
\smartqed  
\usepackage{graphicx}
\usepackage{amsmath,amssymb} 
\usepackage[utf8]{inputenc} 
\usepackage[T1]{fontenc}    
\usepackage{hyperref}       
\usepackage{url}            
\usepackage{booktabs}       
\usepackage{amsfonts}       
\usepackage{nicefrac}       
\usepackage{microtype}      
\usepackage{color}        

\usepackage{times}
\newcommand{\newtext}[1]{\textcolor{black}{{#1}}}

 \hypersetup{ colorlinks, citecolor=green, filecolor=black, linkcolor=blue, urlcolor=blue } 
\def\etal{\emph{et al.}}
\begin{document} 
\title{Image-based  Synthesis for  Deep 3D Human Pose Estimation 
} 


\author{ Gr\'egory~Rogez \and Cordelia Schmid         
}


\institute{
Univ. Grenoble Alpes, Inria, CNRS, Grenoble INP*, LJK, \\38000 Grenoble, France \\
* Institute of Engineering Univ. Grenoble Alpes\\
              \email{gregory.rogez@inria.fr}  
}

\date{Received: date / Accepted: date}

\maketitle

\begin{abstract} 

This paper addresses the problem of 3D human pose estimation in the wild. A significant challenge is the lack of training data, i.e., 2D images of humans annotated with 3D poses. Such data is necessary to train state-of-the-art CNN architectures. Here, we propose a solution to generate a large set of photorealistic synthetic images of humans with 3D pose annotations. We introduce an image-based synthesis engine that artificially augments a dataset of real images with 2D human pose annotations using 3D motion capture data. Given a candidate 3D pose, our algorithm selects for each joint an image whose 2D pose locally matches the projected 3D pose. The selected images are then combined to generate a new synthetic image by stitching local image patches in a kinematically constrained manner. The resulting images are used to train an end-to-end CNN for full-body 3D pose estimation. We cluster the training data into a large number of pose classes and tackle pose estimation as a $K$-way classification problem. Such an approach is viable only with  large training sets such as ours. Our method outperforms most of the published works in terms of 3D pose estimation in controlled environments (Human3.6M) and shows promising results for real-world images (LSP). This demonstrates that CNNs trained on artificial images generalize well to real images. Compared to data generated from more classical rendering engines, our synthetic images do not require any domain adaptation or fine-tuning stage.

\end{abstract}

\vspace{-3mm}

\section{Introduction}

\vspace{-2mm}

Convolutional Neural Networks (CNN) have been very successful for many different tasks in computer vision. However, training these deep architectures requires large scale datasets which are not always available or easily collectable. This is particularly the case for 3D human pose estimation, for which an accurate annotation of 3D articulated poses in large collections of real images is non-trivial:  \newtext{annotating 2D images with 3D pose information is impractical~\cite{bourdev2009poselets} while large scale 3D pose capture is only available in constrained environments through marker-based  (e.g., HumanEva~\cite{SigalBB10}, Human3.6M~\cite{IonescuPOS14}) or  makerless multiview systems (e.g., CMU Panoptic Dataset~\cite{JooLTGNMKNS15}, MARCOnI Dataset~\cite{ElhayekAJTPABST15}). The images captured in such conditions are limited in terms of subjects and environment diversity and do not match well real environments, i.e., real-world scenes with cluttered backgrounds. Moreover, with marker-based systems, the subjects have to wear capture suits with markers on them to which learning algorithms may overfit.} This has limited the development of end-to-end CNN architectures for real-world 3D pose understanding.

\begin{figure*}[ht]
   \centering 
  \includegraphics[width=\textwidth]{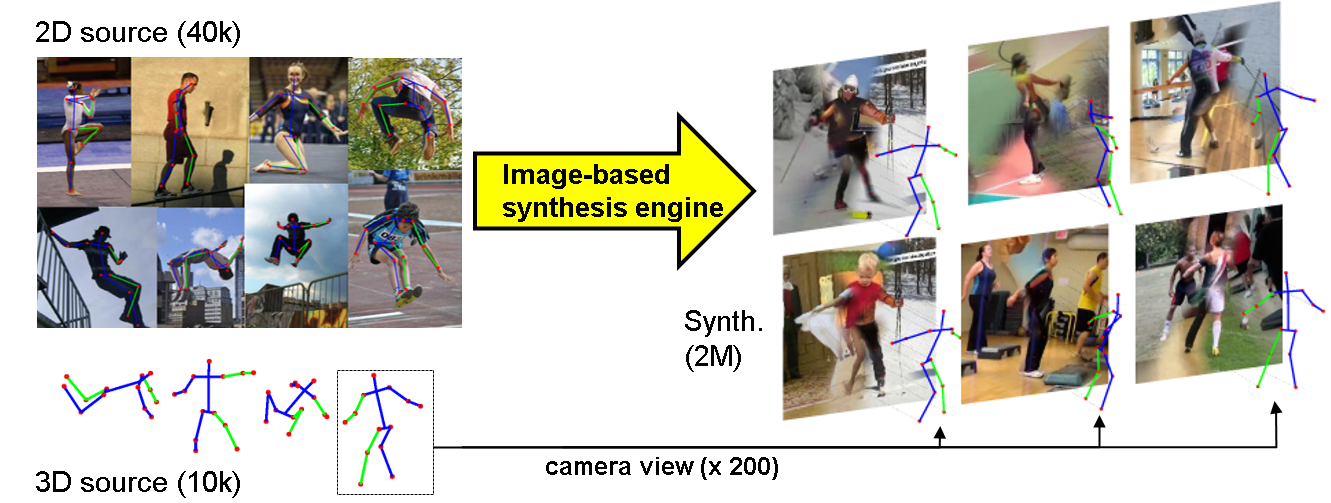}
   \caption{\footnotesize Image-based synthesis engine. Input: real images with manual annotation of 2D poses, and 3D poses
     captured with a motion capture  system.  Output:
     220x220 synthetic images and associated 3D poses.}
     \label{fig:dataset}
 \end{figure*} 
Learning architectures usually augment existing training data by
applying synthetic perturbations to the original images, e.g., jittering exemplars  or applying more complex affine or perspective transformations \cite{JaderbergSZK15}. Such data augmentation has proven to be a crucial stage, especially for training deep architectures. Recent work~\cite{JaderbergSVZ16,PengSAS15,SuQLG15,WuSKYZTX15_ShapeNets} has introduced the use of data synthesis as a solution to train CNNs when only limited data is available. Synthesis can potentially provide infinite training data by rendering 3D CAD models from any camera viewpoint~\cite{PengSAS15,SuQLG15,WuSKYZTX15_ShapeNets}. Fisher et
al.~\cite{DosovitskiyFIHH15_FlowNet} generate a synthetic ``Flying
Chairs'' dataset to learn optical flow with a CNN and show that networks trained on this unrealistic data still generalize very well to existing datasets. In the context of scene text recognition, Jaderberg et al.~\cite{JaderbergSVZ16} trained solely on data produced by a synthetic text generation engine. In this case, the synthetic data is highly realistic and sufficient to replace real data. Although synthesis seems like an appealing solution, there often exists a large domain shift from synthetic to real data~\cite{PengSAS15}. Integrating a human 3D model in a given background in a realistic way is not trivial~\cite{IonescuPOS14}. Rendering a collection of photo-realistic images  (in terms of color, texture, context, shadow) that would cover the variations in pose, body shape, clothing and scenes is a challenging task.  

Instead of rendering a human 3D model, we propose an image-based synthesis approach that makes use of motion capture data to augment an existing dataset of real images with 2D pose annotations. Our system synthesizes a very large number of new images showing more pose configurations and, importantly,
it provides the corresponding 3D pose annotations (see Figure~\ref{fig:dataset}).
For each candidate 3D pose in the motion capture library, our system combines several annotated images to generate a synthetic image of a human in this particular pose. This is achieved  by ``copy-pasting'' the image information corresponding to each joint in a kinematically constrained manner.
Given this large ``in-the-wild'' dataset, we implement an end-to-end CNN architecture for 3D pose estimation. Our approach first clusters the 3D poses into $K$ pose classes. Then, a $K$-way CNN classifier is trained to return a distribution over probable pose classes given a bounding box around the human in the image. Our method outperforms most state-of-the-art results in terms of 3D pose estimation in controlled environments and shows promising results on images captured ``in-the-wild''.  The work presented in this paper is an extension of \cite{RogezS16}.  \newtext{We provide an additional comparison of our image-based synthesis engine with a more classical approach based on rendering a human 3D model. The better performance of our method shows that for training a deep pipeline with a classification or a regression objective, it is more important to produce locally photorealistic data than globally coherent data.}

\begin{figure*}[t]
  \centering
   \includegraphics[width=\textwidth]{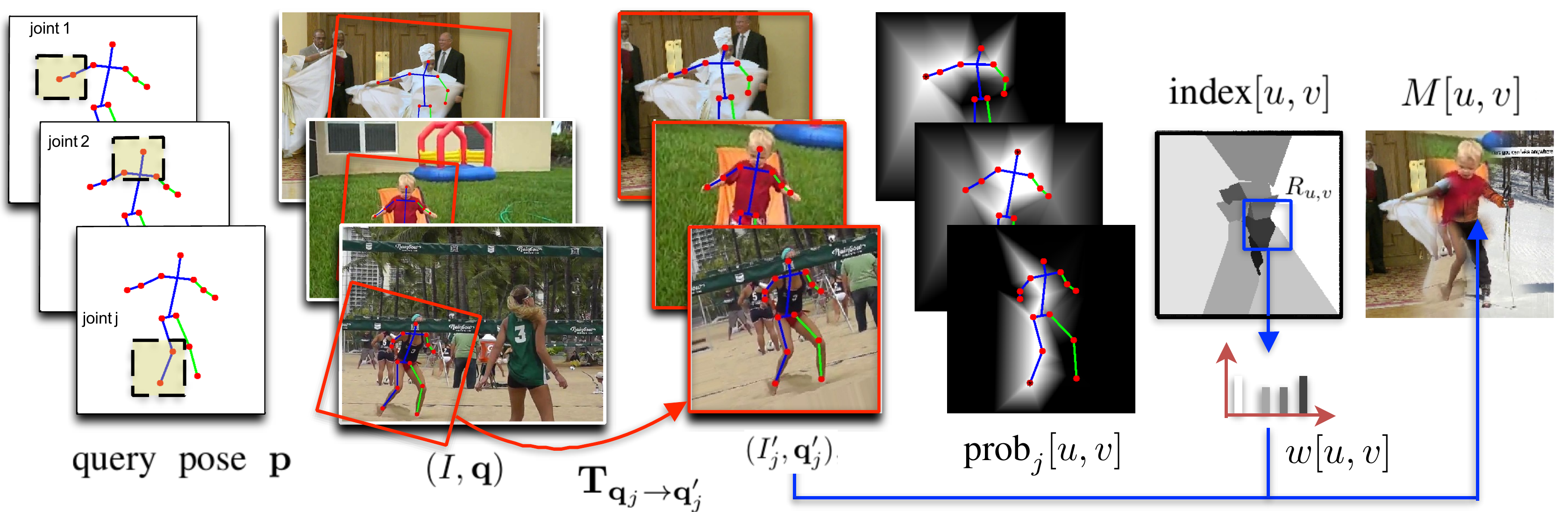}
 \caption{\footnotesize Synthesis engine. From left to right:  for each joint $j$ of a 2D query pose ${{\bf p}}$ (centered in a  $220 \times 220$ bounding box), we align all the annotated 2D poses w.r.t the limb and search for the best pose match, resulting in a list of $n$ matches $\{(I_j',    {    \bf q}_j'), j=1...n\}$ where $I_j'$ is obtained after transforming $I_j$ with ${{\bf T}{{\bf q}_j \rightarrow {\bf q}_j'}}$. For each retrieved pair, we compute a probability map $\textrm{prob}_j[u,v]$. These $n$ maps are used to compute  $\text{index}[u,v] \in \{1...n\}$, pointing to the image $I_{j}'$ that should be used for a particular pixel $(u,v)$. Finally, our blending algorithm computes each pixel value of the synthetic image $M[u,v]$ as the weighted sum over all aligned images $I_j'$, the weights being calculated using a histogram of indexes in a squared region $R_{u,v}$ around  $(u,v)$.} 
      \label{fig:mosaique}
\end{figure*}
\subsection{Related work}

 {\bf 3D human pose estimation in monocular images.} 
Recent approaches employ CNNs for 3D pose estimation in monocular
images~\cite{ChenWLSWTLCC16,LiZC15,PavlakosZDD17} or in videos~\cite{ZhouZLDD16}. Due to the
lack of large scale training data, they are usually trained  (and
tested) on 3D motion capture data in constrained
environments~\cite{LiZC15}. Pose understanding in natural images
is usually limited to 2D pose
estimation~\cite{ChenY14,TompsonJLB14,ToshevS14_DeepPose}. \newtext{Motivated by these well-working off-the-shelf 2D detectors and inspired by earlier work in single view 3D pose reconstruction~\cite{Mori06,RamakrishnaKS12,SigalB06,Taylor00}}, recent work
also tackles 3D pose understanding from 2D
poses~\cite{AkhterB15,ChenR17,FanZZW14,Moreno17,TomeRA17}. Some
approaches use as input the 2D joints automatically provided by a 2D
pose detector~\cite{ChenR17,Moreno17,Simo-SerraRATM12,WangWLYG14}, while others jointly
solve  the  2D and 3D pose estimation~\cite{Simo-SerraQTM13,TomeRA17,ZhouT14}. 
\newtext{Most similar to ours are the architectures that take advantage of the different sources of training data, i.e., indoor images with motion capture 3D poses and real-world images with 2D annotations~\cite{MehtaRCFSXT17,IqbalGG16,ZhouHSXW17}. Iqbal et al.~\cite{IqbalGG16} use a dual-source approach that combines 2D pose estimation with 3D pose retrieval. 
Mehta et al.~\cite{MehtaRCFSXT17} propose a 2D-to-3D knowledge transfer to generalize to in-the-wild images, using pre-trained 2D pose networks to initialize the 3D pose regression networks. The architecture of~\cite{ZhouHSXW17} shares the common representations between the 2D and the 3D tasks.}
Our method uses the same two training sources, i.e., images with annotated 2D pose and 3D motion capture data. However, we
combine both sources off-line to generate a large training set that is
used to train an end-to-end CNN 3D pose classifier. This is shown to
improve over~\cite{IqbalGG16}, which can be explained by the fact that
training is performed in an end-to-end fashion. 

\noindent {\bf Synthetic pose data.} A number of works have considered the use of synthetic data for human pose estimation. Synthetic data have been used for upper body~\cite{ShakhnarovichVD03}, full-body silhouettes~\cite{DAgarwalT06}, hand-object interactions~\cite{romero_hands_2010}, full-body pose from depth~\cite{ShottonFCSFMKB11} or egocentric RGB-D scenes~\cite{RogezSR15}. Zuffi and Black~\cite{ZuffiB15} used a 3D mesh-model to sample synthetic exemplars and fit 3D scans. Recently, Chen et al.~\cite{ChenWLSWTLCC16} trained a human 3D pose regressor on synthetic training images rendered from such a 3D mesh-model. Similarly, \cite{HuangR17} trained a human detector for unusual pedestrian using synthetic data generated by a game engine. In both cases, a domain adaptation stage was necessary to generalize to real images. In~\cite{HattoriBKK15}, a scene-specific pedestrian detector was learned  without real data while~\cite{EnzweilerG08} synthesized  virtual samples with a generative model to enhance the classification performance of a discriminative model. In~\cite{HornungDK07},  pictures of 2D characters were animated by fitting and deforming a 3D mesh model. Later, \cite{PishchulinJATS12} augmented labelled training images with small perturbations in a similar way. These methods require a perfect segmentation of the humans in the images. Park and Ramanan~\cite{ParkR15}  synthesized hypothetical poses for tracking by applying geometric transformations to the first frame of a video sequence. We also use  image-based synthesis to generate images but our rendering engine combines image regions from several images to create  images with associated 3D poses.


 \section{Image-based synthesis engine}
  
 At the heart of our approach is an image-based synthesis engine that
 artificially generates ``in-the-wild'' images  with 3D pose
 annotations. Our method takes as input a dataset of real images with 2D
 annotations and a library of 3D motion capture
 data, and generates a large number of synthetic images with associated
 3D poses (see Figure~\ref{fig:dataset}). 
We introduce an image-based rendering engine that augments the
existing database of annotated images with a very large set of
photorealistic images covering more body pose configurations
than the original set.  
\newtext{This is done by projecting the motion capture 3D poses on random camera views to obtain a set of 2D poses for which new images are synthesized by selecting and stitching image patches in a kinematically constrained manner. }
Our synthesis process consists
 of two stages:  the mosaic construction stage that selects and stitches image
 patches together and a pose-aware blending  process that improves image quality and erases  patch seams. These are discussed in the following subsections. 
Figure~\ref{fig:mosaique} summarizes our synthesis process.

 \subsection{Motion capture guided image mosaicing}

\begin{figure}[h]
   \centering 
  \includegraphics[width=\columnwidth]{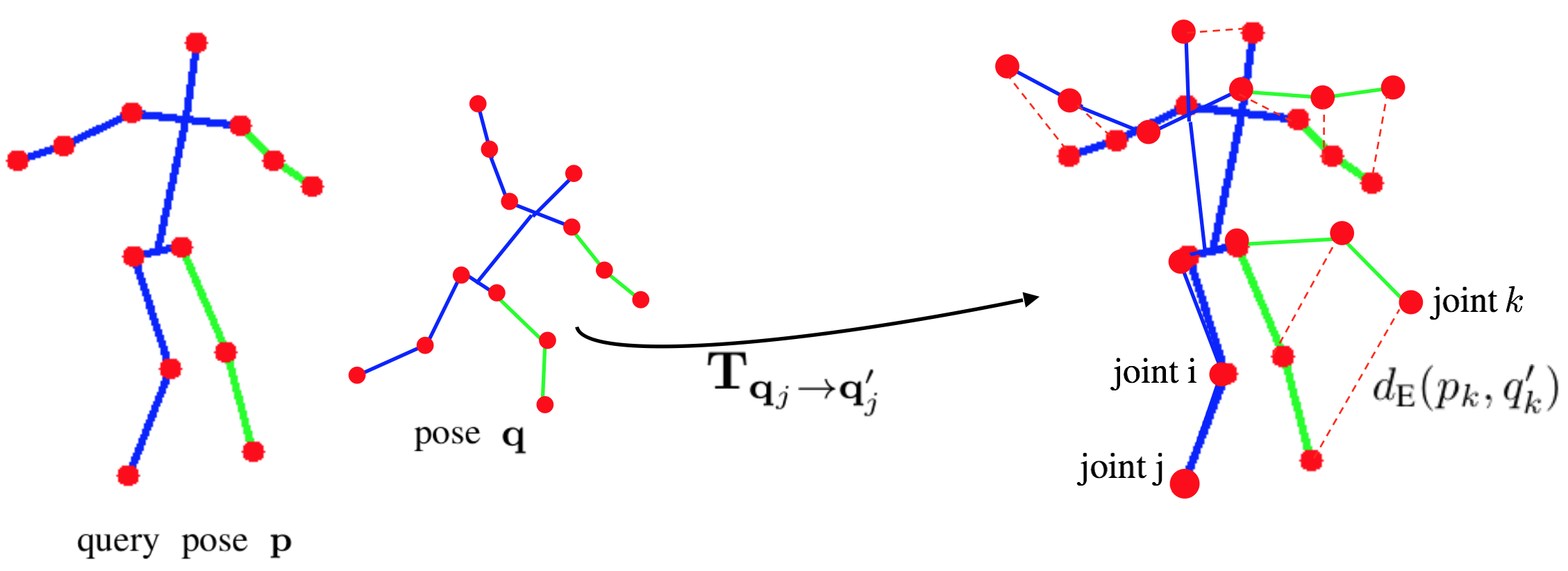}
  \caption{\footnotesize \newtext{Illustration of the cost function employed to find pose matches. We show two poses aligned at joint $j$ with red lines across all the other joints denoting contributors to the distance.} }
     \label{fig:distance}
 \end{figure}

Given a 3D pose with $n$ joints  ${\bf P }\in \mathbb{R}^{n \times
    3}$, and its projected 2D joints ${{\bf p}=\{p_{j} , j=1...n\}}$  in  a particular camera view, we want to find for each joint $j \in \{1...n \}$ an image whose annotated 2D pose presents a similar kinematic configuration around $j$. To do so, we define a distance function between 2 different 2D poses ${\bf p}$ and ${\bf q}$, conditioned on joint $j$ as:
\begin{align}
  D_{j}({\bf p}, {\bf q})= \sum_{k=1}^{n} d_{\text{E}}(p_k, q'_k)  \label{eq:posedist} 
\end{align}
where $d_{\text{E}}$ is the Euclidean distance. ${\bf q'}$ is the
aligned version of ${\bf q}$ with respect to joint $j$ after applying
a rigid transformation ${{\bf T}_{{\bf q}_j \rightarrow {\bf q}_j'}}$,
which respects $q_{j}'=p_{j}$ and $q_{i}'=p_{i}$ , where $i$ is the
farthest directly connected joint to $j$ in ${\bf p}$.  \newtext{The rigid transformation ${{\bf T}_{{\bf q}_j \rightarrow {\bf q}_j'}}$ is obtained by combining the translation ${{\bf t}_{{ q}_j \rightarrow {p}_j}}$ aligning joint $j$ in ${\bf p}$ and ${\bf q}$, and the rotation matrix ${{\bf R}_{{ q}_i \rightarrow {p}_i}}$ aligning joint $i$ in ${\bf p}$ and ${\bf q}$:
\begin{align}
 {{\bf T}_{{\bf q}_j \rightarrow {\bf q}_j'}}(p_k)= {{\bf R}_{{ q}_i \rightarrow {p}_i}} (p_k)+{{\bf t}_{{ q}_j \rightarrow {p}_j}}. \label{eq:transf} 
\end{align}
An example of such transformation is given in Fig.~\ref{fig:distance}.}

The function $D_{j}$ measures the similarity between 2 joints by aligning and taking into account the entire poses.  
To increase the influence of neighboring joints, we weight the distances $d_{\text{E}}$ between each pair of joints $\{(p_k, q'_k), k=1...n\}$  according to their distance to the query joint $j$ in both poses. Eq.~\ref{eq:posedist} becomes:
 \begin{align}
   D_{j}({\bf p}, {\bf q})= \sum_{k=1}^{n} (w_k^j({\bf p}) + w_k^j({\bf q})) \;d_{\text{E}}(p_k, q'_k) \label{eq:posesim} 
 \end{align}
where weight $w_k^j$ is inversely proportional to the distance between joint $k$ and the query joint $j$, i.e., $w_k^j({\bf p})=1/d_{\text{E}}(p_k, p_j)$ and normalized so that $\sum_k w_k^j({\bf p})=1$. This cost function is illustrated in Fig.~\ref{fig:distance}.

\begin{figure*}[ht]
   \centering 
  \includegraphics[width=\textwidth]{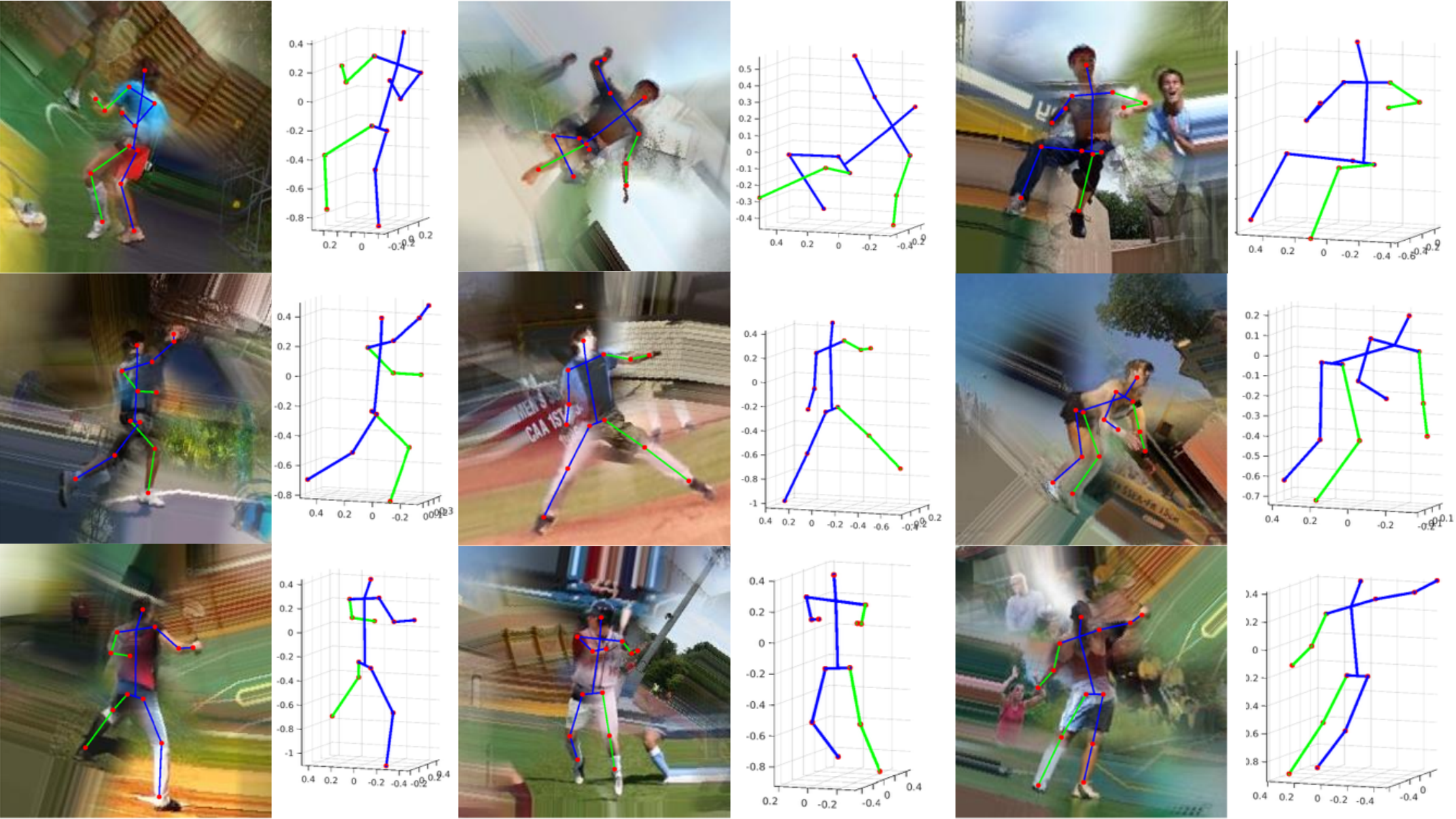}
  \caption{\footnotesize \newtext{Examples of synthetics images generated using  Leeds Sport dataset
(LSP)\cite{JohnsonE10} and CMU motion capture
dataset  as 2D and 3D sources respectively. For each case, we show the 2D pose overlaid on the image  and the corresponding orientated 3D pose.}}
     \label{fig:qualitative_results}
 \end{figure*}
For each joint $j$ of the query pose  ${{\bf
    p}}$,
we retrieve from our dataset  $\mathbb{Q}=\{(I_1,{\bf q}_1)\dots(I_N,{\bf q}_N)\}$ of images and annotated 2D poses: 
 \begin{align}
  {\bf q}_j = \text{argmin}_{{\bf q} \in \mathbb{Q}} D_{j}({\bf p}, {\bf q})  \quad \forall j  \;  \in \{1...n\}. \label{eq:posesearch} 
 \end{align}
 \newtext{In practice, we do not search for self-occluded joints, i.e., joints occluded by another body part, that can be labelled as such by simple 3D reasoning.}
We obtain a list of $n$ matches $\{(I_j',{\bf q}_j'), j=1...n\}$ where
$I_j'$ is the cropped image obtained after transforming $I_j$ with
${{\bf T}_{{\bf q}_j \rightarrow {\bf q}_j'}}$. Note that a same pair
$(I,{\bf q})$ can appear multiple times in the list of candidates,
i.e., being a good match for several joints.

Finally, to render a new image, we need to select the candidate images $I_j'$ to be used for each pixel $(u,v)$. Instead of using regular patches, we compute a probability map $p_j[u,v]$ associated with each pair $(I_j',{\bf q}_j')$ based on local matches measured by $d_{\text{E}}(p_k, q'_k)$ in Eq.~\ref{eq:posedist}. To do so, we first apply a Delaunay triangulation to the set of 2D joints in $\{{\bf q}_j'\}$  obtaining a partition of the image into triangles, according to the selected pose. Then, we assign the probability $\textrm{prob}_j(q'_k)=\textrm{exp}(-d_{\text{E}}(p_k, q'_k)^2/\sigma^2) $ to each vertex $q'_k$. We finally compute a probability map $\textrm{prob}_j[u,v]$ by interpolating values from these vertices using barycentric interpolation inside each triangle.
The resulting $n$ probability maps are concatenated and  an index map $\text{index}[u,v] \in \{1...n\}$ can be computed as follows: 
\begin{align}
\text{index}[u,v]=\text{argmax}_{j \in \{1 \dots n\}} \; \textrm{prob}_j[u,v], 
\end{align}
this map pointing to the training image $I_{j}'$ that should be used for each pixel $(u,v)$. A mosaic $M[u,v]$ can be generated by ``copy-pasting'' image information at pixel $(u,v)$ indicated by $\text{index}[u,v]$:
\begin{align}
M[u,v]=I_{j^*}'[u,v] \quad \text{with} \quad j^*=\text{index}[u,v]. 
\end{align}

 \subsection{Pose-aware image blending}
 
The mosaic $M[u,v]$ resulting from the previous stage presents significant artifacts at the boundaries between image regions. Smoothing is necessary to prevent the learning algorithm from interpreting these artifacts as discriminative pose-related features. We first experimented with off-the-shelf image filtering and alpha blending algorithms, but the results were not satisfactory. Instead, we propose a new pose-aware blending algorithm that maintains image information on the human body while erasing most of the stitching artifacts. 
For each pixel $(u,v)$, we select a surrounding squared region $R_{u,v}$ whose size varies with the distance $d_{u,v}$ of pixel $(u,v)$ to the pose:  \newtext{
 \begin{align}
  R_{u,v}=\alpha + \beta d_{u,v}.
 \end{align}
$R_{u,v}$ will be larger when far from the body and smaller nearby. The distance $d_{u,v}$ is computed using a distance transform to the rasterisation of the 2D skeleton. In this paper, we empirically set $\alpha$=6 pixels and $\beta$=0.25 to synthesise  220$\times$220 images.}
 Then, we evaluate how much each image $I_{j}'$ should contribute to the value of pixel $(u,v)$ by building a histogram of the image indexes inside the region $R_{u,v}$:
 \begin{align}
   w[u,v]=\text{Hist}(\text{index}(R_{u,v})),  
 \end{align}
 where the weights are normalized so that $\sum_j w_j[u,v] =1$. The final mosaic $M[u,v]$ (see examples in Figure~\ref{fig:dataset}) is then computed as the weighted sum over all aligned images:
 \begin{align}
   M[u,v] = \sum_j w_j[u,v] I_j'[u,v].
 \end{align}
This procedure produces plausible images that are kinematically correct and locally photorealistic. See examples presented in Figure~\ref{fig:dataset} and Figure~\ref{fig:qualitative_results}.

 \section{CNN for full-body 3D pose estimation}

\begin{figure*}[thb]
   \centering 
  \includegraphics[width=\textwidth]{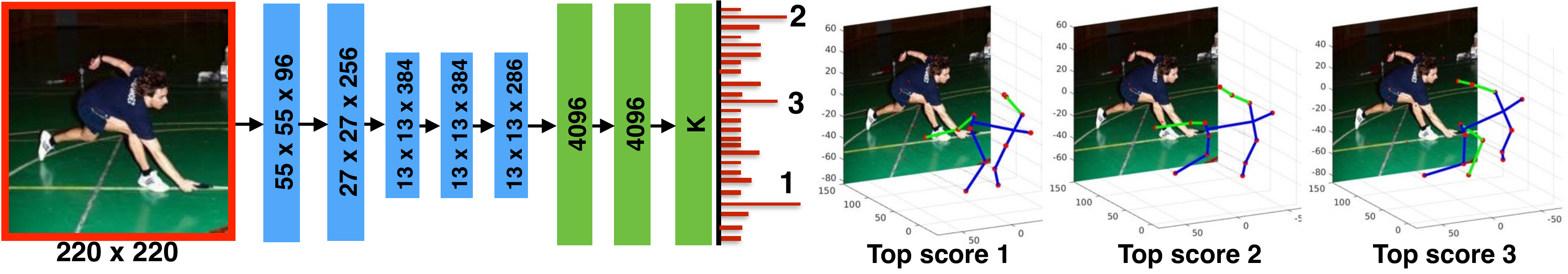}
\caption{\footnotesize  CNN-based pose classifier. We show the different layers and their corresponding dimensions, with convolutional layers depicted in blue and fully connected ones in green.  
The output is a distribution over K pose classes. Pose estimation is obtained by taking the highest score in this distribution. We show on the right the 3D poses for 3 highest scores. For this example, the top scoring class (top score 1) is correct. }
     \label{fig:CNN}
 \end{figure*}

Human pose estimation has been addressed as a classification problem in the past~\cite{BissaccoYS06,OkadaS08,RogezROT12,RogezSR15}. Here, the 3D pose space is partitioned into $K$ clusters and a $K$-way classifier is trained to return a distribution over pose classes. Such a classification approach allows modeling multimodal outputs in ambiguous cases, and produces multiple hypothesis that can be rescored, e.g., using temporal information. Training such a classifier requires a reasonable amount of data per class which implies a well-defined and limited pose space (e.g. walking action)~\cite{BissaccoYS06,RogezROT12}, a large-scale synthetic dataset~\cite{RogezSR15} or both~\cite{OkadaS08}.  
Here, we introduce a CNN-based classification approach for full-body 3D pose estimation. Inspired by the DeepPose algorithm~\cite{ToshevS14_DeepPose} where the AlexNet CNN architecture \cite{NIPS2012_Alexnet} is used for full-body 2D pose regression, we select the same architecture and adapt it to the task of 3D body pose classification. This is done by adapting the last fully-connected layer to output a distribution of scores over pose classes as illustrated in Figure~\ref{fig:CNN}. 
 Training such a classifier requires a large amount of training data that we generate using our image-based synthesis engine. 

Given a library of motion capture data and a set of camera views, we synthesize for each 3D pose a $220\times220$ image. This size has proved to be adequate for full-body pose estimation~\cite{ToshevS14_DeepPose}. The 3D poses are then aligned with respect to the camera center and translated to the center of the torso, i.e., the average position between shoulders and hips coordinates. 
In that way, we obtain orientated 3D poses that also contain the viewpoint information.  
 We cluster the resulting 3D poses to define our classes which will correspond to groups of similar orientated 3D poses, i.e., body pose configuration and camera viewpoint.
 We empirically found that $K$=5000 clusters was a sufficient number of clusters and that adding more clusters did not further improve the results. For evaluation, we return the average 2D and 3D poses of the top scoring class. 
    \begin{table*}[htb]
\caption{Impact of synthetic data on the performances for the regressor and the classifier. The 3D pose estimation results are given following the protocol P1 of Human3.6M (see text for details).}
\centering
\begin{tabular}{l|ccccc }
\hline
  Method & Type & 2D source    & 3D source & \newtext{ Training } & Error    \\ 
   & of images  & size&  size  &  \newtext{pairs }&   (mm) \\  
\hline 
Reg. & Real &  17,000 &  17,000  &  17,000 & 112.9   \\ 
Class. & Real &    17,000 &  17,000 &  17,000  &149.7   \\ 
\hline
Reg. & Synth &  17,000 & 190,000 & 190,000  & 101.9 \\ 
Class. & Synth &  17,000   & 190,000 & 190,000 &  97.2    \\ 
\hline
Reg. & Real & 190,000   & 190,000  & 190,000 &139.6    \\ 
Class. & Real & 190,000   & 190,000& 190,000   &97.7  \\ 
\hline
Reg. & Synth + Real &  207,000 & 190,000& 380,000  &125.5  \\ 
Class. & Synth + Real & 207,000  & 190,000& 380,000  &\bf{88.1}   \\ 
\hline
\end{tabular} 
\label{tab:H36M}
\end{table*}

To compare with~\cite{ToshevS14_DeepPose}, we also train a holistic pose regressor, which regresses to 2D and 3D poses (not only 2D). To do so, we concatenate the 3D coordinates expressed in meters normalized to the range $[-1,1]$,  with the 2D pose coordinates, also normalized in the range $[-1,1]$ following~\cite{ToshevS14_DeepPose}.

 \section{Experiments}

We address 3D pose estimation in the wild. However, there does not
exist a dataset of real-world images with 3D annotations. We thus
evaluate our method in two different settings using existing datasets:
(1) we validate our 3D pose predictions using 
Human3.6M~\cite{IonescuPOS14} which provides accurate 3D and 2D poses for 15
different actions captured in a controlled indoor
environment; (2) we evaluate on the Leeds Sport dataset
(LSP)~\cite{JohnsonE10} that presents real-world images together with full-body 2D pose annotations.  We demonstrate competitive results
with state-of-the-art methods for both of them. 

Our image-based rendering engine requires two different training
sources: 1) a 2D source of images with 2D pose annotations and 2) a motion capture 3D source. We consider two different datasets for each: for 3D
poses we use the CMU motion capture
dataset\footnote{http://mocap.cs.cmu.edu} and the Human3.6M 3D
poses~\cite{IonescuPOS14}, and for 2D pose annotations the
MPII-LSP-extended dataset~\cite{PishchulinITAAG15} and the Human3.6M 2D poses and images.

{\noindent \bf Motion capture 3D source.} The CMU motion capture dataset consists of 2500 sequences and a total of 140,000 3D poses. We align the 3D poses w.r.t. the torso and select a subset of 12,000 poses, ensuring that selected poses have at least one joint 5 cm apart. In that way, we densely populate our pose space and avoid repeating common poses, e.g., neutral standing or walking poses which are over-represented in the dataset. For each of the 12,000 original motion capture poses, we sample 180 random virtual views with azimuth angle spanning 360 degrees and elevation angles in the range $[-45,45]$. We generate over 2 million pairs of 3D/2D pose configurations (articulated poses + camera position and angle). 
For Human3.6M, we  
randomly selected a subset of 190,000 orientated 3D poses, discarding similar poses, i.e., when the average Euclidean distance of the joints is less than 15mm as in~\cite{IqbalGG16}.

 \begin{figure*}[t]
   \centering 
  \includegraphics[width=0.85\textwidth]{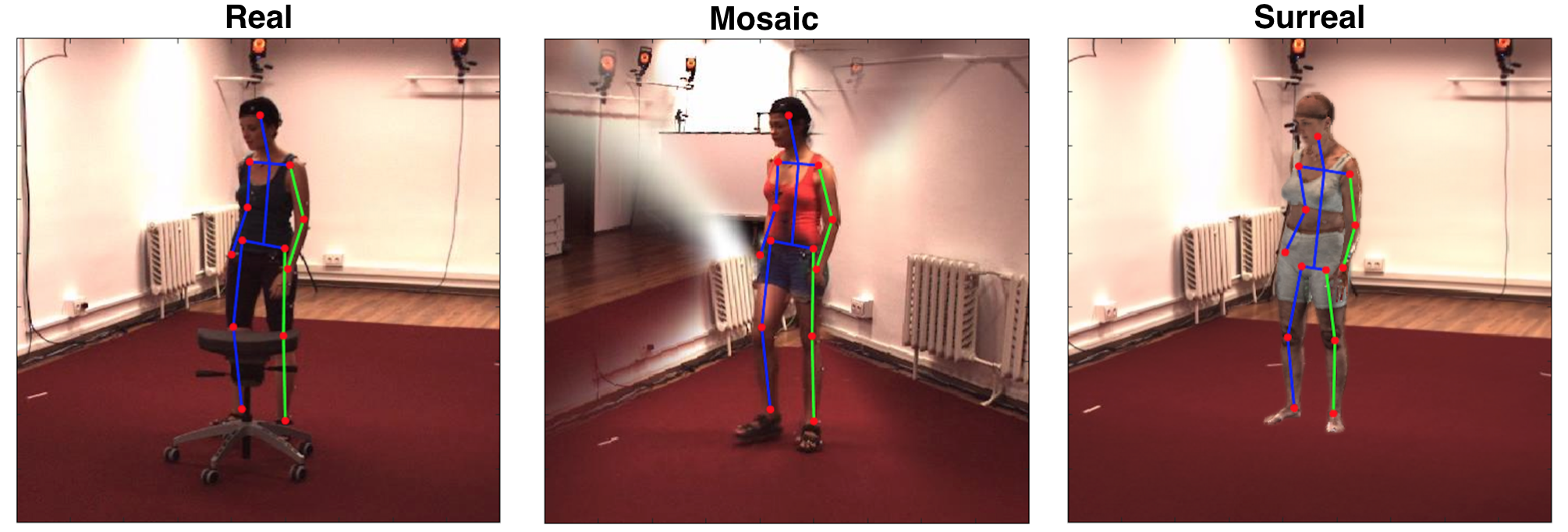}
\caption{\footnotesize  Human 3.6M real and synthetic data. We show on the left a training image from protocol 2 with the overlayed 2D pose. In the middle, we show a ``mosaic'' image synthetized using the 2D pose from the real image on the left.  In this case, the mosaic has been built by stitching image patches from 3 different subjects. On the right, we show a synthetic ``surreal'' image obtained after rendering the SMPL model using the 3D pose from the real image on the left. Note that for more realism, the surreal image is rendered at the exact same 3D location in the motion capture room, using the same camera and background as in the real image. }
     \label{fig:surreal}
 \end{figure*} 
{\noindent \bf 2D source.}  For the training dataset of real
images with 2D pose annotations, we use the MPII-LSP-extended
\cite{PishchulinITAAG15} which is a concatenation of the extended LSP
\cite{JohnsonE11} and the MPII dataset \cite{andriluka14cvpr}. Some of
the poses were manually corrected as a non-negligible number of
annotations are not accurate enough or completely wrong (eg., right-left inversions or bad ordering of the joints along a limb).  
 We mirror the images to double the size of the training set,
 obtaining a total of 80,000 images with 2D pose annotations.
For Human3.6M, we consider  
the 4 cameras and  create  a pool of 17,000
images and associated 2D poses that we also mirror. \newtext{To create a diverse set of images, we ensure that the maximum joint-to-joint distance between 
 corresponding 3D poses is over 5 cm, i.e., similar poses have at least one joint 5 cm apart in 3D. }

 \subsection{Evaluation on Human3.6M Dataset}

To compare our results with  recent work in 3D pose estimation~\cite{IqbalGG16}, we follow the protocol introduced in \cite{KostrikovG14} and employed in \cite{IqbalGG16}: we consider six subjects (S1, S5, S6, S7, S8 and S9) for training, use every $64^{th}$ frame of subject S11 for testing and evaluate the 3D pose error (mm) averaged over the 13 joints. We refer to this protocol by P1. As in~\cite{IqbalGG16}, we measure a 3D pose error that aligns the  pose by a rigid transformation, but we also report the absolute error.  

We first evaluate the impact of our synthetic data on the performances
for both the regressor and classifier. The results are reported in
Table~\ref{tab:H36M}. We can observe that when considering few training
images (17,000), the regressor clearly outperforms the classifier
which, in turns, reaches better performances when trained on larger
sets. This can be explained by  
the fact that the classification approach requires a sufficient amount of examples.
 We, then, compare results when training both regressor and classifier
 on the same 190,000 poses considering a)~synthetic data generated
 from Human3.6M, b)~the real images corresponding to the 190,000 poses and
 c)~the synthetic and real images together. We observe that the
 classifier has similar performance when trained on synthetic or
 real images, which means that our image-based rendering engine
 synthesizes useful data. Furthermore, we can see that the classifier
 performs much better  when trained on synthetic and real images together. This means that our data  is different from
 the original data and allows the classifier to learn better
 features. Note that we retrain AlexNet from scratch. We found that it performed better than just fine-tuning a model pre-trained on Imagenet (3D error of 88.1mm vs 98.3mm with fine-tuning).
 
 In Table~\ref{tab:H36M_sota}, we compare our results to three state-of-the-art approaches. Our best classifier, trained with a combination of synthetic and real data, outperforms these methods in terms of 3D pose estimation by a margin. Note that even though we compute 3D pose error after 3D alignment, our method initially estimates absolute pose (with orientation w.r.t. the camera). That is not the case of Bo et al.~\cite{BoS10} for instance, who estimate a relative pose and do not provide 3D orientation.
 \begin{table} 
\caption{Comparison with state-of-the-art methods on Human3.6M following protocol P1 that measures an aligned 3D pose distance.}
\centering
\begin{tabular}{c|ccc}
\hline 
  Method & 2D source  & 3D source  &   Error   \\ 
     &  size &  size &  (mm) \\ 
\hline
Bo\&Sminchisescu~\cite{BoS10} & 120,000 & 120,000  &117.9 \\
Kostrikov\&Gall~\cite{KostrikovG14}&  120,000 & 120,000  &115.7 \\ 
Iqbal et al.~\cite{IqbalGG16} & 300,000  & 380,000 & 108.3 \\ 
Ours  &207,000 & 190,000 & \bf{88.1} \\
\hline
\end{tabular} 
\label{tab:H36M_sota}
\end{table}

\noindent {\bf Comparison with classical synthetic images.} We make additional experiments to further understand how useful our data is with respect to more classical synthetic data, i.e. obtained by rendering a human 3D model as in~\cite{ChenWLSWTLCC16,VarolRMMBLS17}.  To do so, we consider the same 190,000 poses from the previous experiments and render the SMPL 3D human mesh model~\cite{SMPL:2015} in these exact same poses using the body parameters and texture maps from~\cite{VarolRMMBLS17}. To disambiguate the type of synthetic data, we refer to these new rendered images as ``surreal'' images, as named in~\cite{VarolRMMBLS17} and refer to our data as ``mosaic'' images. Note that for more realism and to allow for a better comparison, we place the 3D model in the exact same location within the Human3.6M capture room and use the backgrounds and camera parameters of the corresponding views to render the scenes. An example of the resulting surreal images is visualized in Figure~\ref{fig:surreal} where we also show the corresponding original real image as well as our mosaic image obtained for the exact same pose. When the 2D annotations are accurate and consistent, as it is the case with the Human3.6M dataset, our algorithm produces very plausible images that are locally photorealistic and kinematically correct without significant artefacts at the boundaries between the image patches. Note that for this experiment, we use the poses and images from subjects S1, S5, S6, S7 and S8 to generate our synthetic sets, i.e. removing S9 from the training set. This allows us to also evaluate on a second protocol (P2) employed in~\cite{LiZC15,TekinRLF16,ZhouZLDD16} where only these 5 subjects are used for training.

   \begin{figure*}[t]
   \centering 
  \includegraphics[width=\textwidth]{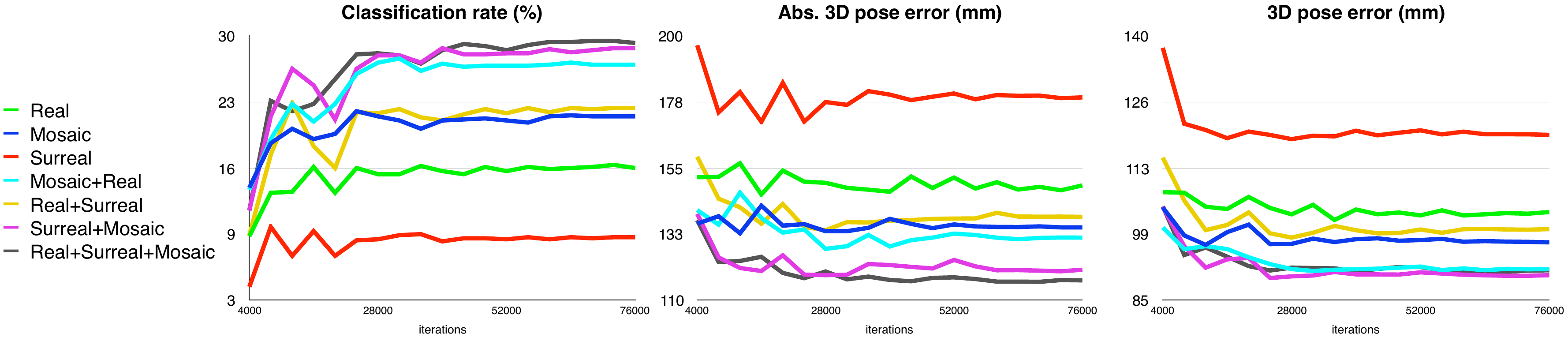}
\caption{\footnotesize  Human pose estimation results on Human3.6M when training the classifier with different combinations of real, surreal and mosaic data. We report classification rate, absolute 3D pose estimation error and 3D pose error after rigid alignment.  We show the performance with respect to the number of training iterations. The three sets of training images all correspond to the same 190,000 3D poses and are labelled with the same class labels.}
     \label{fig:numres}
 \end{figure*} 
\begin{table*}[ht]
\resizebox{\linewidth}{!}{
\begin{tabular}{lc|ccccccccc}
\hline
Method & Im  & Directions & Discussion & Eat & Greet & Phone & Pose & Purchase & Sit & SitDown \\
\hline 
Tekin~\etal \cite{TekinRLF16} &   & 102.4 & 147.7 & 88.8 & 125.3 & 118.0 & 112.3 & 129.2 & 138.9 & 224.9 \\
Zhou~\etal  \cite{ZhouZLDD16}&   & 87.4 & 109.3 & 87.0 & 103.2 & 116.2 & 106.9 & 99.8 & 124.5 & 199.2 \\
Du~\etal \cite{DWLHGWKG16}&    & 85.1 & 112.7 & 104.9 & 122.1 & 139.1 & 105.9 & 166.2 & 117.5 & 226.9 \\
\hline 
Li~\etal \cite{LiZC15}& \checkmark    & - & 134.1 & 97.4 & 122.3 & - & - & - & - & - \\
Li~\etal \cite{LiZC16}& \checkmark    & - & 133.5 & 97.6 & 120.4 & - & - & - & - & - \\ 
Tekin~\etal \cite{TekinKSLF16} & \checkmark    & - & 129.1 & 91.4 & 121.7 & - & - & - & - & - \\
Sanzari~\etal \cite{SanzariNP16} & \checkmark   & {\bf 48.8} & {\bf 56.3} & 96.0 &84.8& 96.5 &  66.3 & 107.4 & 116.9 & 129.6\\ 
Zhou~\etal  \cite{ZSZLW16}& \checkmark   & 91.8 & 102.4 & 97.0 & 98.8 & 113.4 & 90.0 & 93.8 &  132.2 & 159.0 \\ 
Tome~\etal  \cite{TomeRA17}& \checkmark    & 65.0 &73.5 & 76.8 & 86.4 & 86.3 &68.9& 74.8&110.2 & 173.9\\
Moreno~ \cite{Moreno17}& \checkmark    & 69.5 &80.2 & 78.2 & 87.0 & 100.8 &76.0& 69.7&104.7 & 113.9\\
Rogez~\etal \cite{RogezWS17}& \checkmark  &  76.2 &  80.2 &  75.8 & 83.3&  92.2 &79.0 & 71.7 &  105.9 &  \bf 127.1 \\
Pavlakos~\etal \cite{PavlakosZDD17}&  \checkmark    & 58.6 &64.6 & {\bf 63.7} & {\bf 62.4} & {\bf 66.9 }&{\bf  57.7}& {\bf 62.5}&{\bf 78.6} & {\bf  103.5}\\
\hline
Ours (real+mosaic)& \checkmark   & 94.5  & 110.4 & 109.3 & 143.9 & 125.9 & 95.5 & 89.8 & 134.2 & 179.2 \\
Ours (real+mosaic+surreal)& \checkmark   & 87.7 & 100.7 & 93.6 & 139.6 & 107.9 &88.1 & 78.9 &119.0& 171.9\\
\hline
\hline
Method & Im  & Smoke &   Photo & Wait & Walk  & WalkDog & WalkTogether & & Avg. (All) & Avg. (6) \\
\hline 
Tekin~\etal \cite{TekinRLF16} &   & 118.4 & 182.7 & 138.7 & {\bf 55.1 }& 126.3 &65.8 & & 125.0 & 121.0 \\
Zhou~\etal \cite{ZhouZLDD16}&    & 107.4 & 143.3 & 118.1 & 79.4 & 114.2 & 97.7 & & 113.0 & 106.1 \\
Du~\etal \cite{DWLHGWKG16}&   & 120.0 & 135.9 & 117.6 & 99.3 & 137.4 & 106.5 & & 126.5 & 118.7 \\
\hline 
Li~\etal \cite{LiZC15}& \checkmark    & - & 166.2 & - & 68.5 & 132.5 & - & & - & 121.3 \\
Li~\etal \cite{LiZC16}& \checkmark    & - & 163.3 & - & 73.7 & 135.2 & - & & - & 121.6 \\ 
Tekin~\etal \cite{TekinKSLF16} & \checkmark  & - & 162.2 & - & 65.7 & 130.5 & - & & - & 116.8 \\
Sanzari~\etal \cite{SanzariNP16}  & \checkmark   &97.8 &  105.6 & 65.9 & 92.6 & 130.5 & 102.2 & &93.1 & -\\
Zhou~\etal  \cite{ZSZLW16}& \checkmark    & 106.9 & 125.2 & 94.4 & 79.0 & 126.0& 99.0 &   & 107.3 & - \\ 
Tome~\etal  \cite{TomeRA17}&  \checkmark   &85.0 &110.7& 85.8& 71.4 & 86.3& 73.1 & &88.4 & -\\
Moreno~ \cite{Moreno17}&  \checkmark   &89.7 &102.7& 98.5& 79.2 & 82.4& 77.2 & &87.3 & -\\
Rogez~\etal \cite{RogezWS17} & \checkmark  &   88.0 & 105.7 & 83.7 & 64.9 & 86.6& 84.0 &   & 87.7& 83.0 \\
Pavlakos~\etal \cite{PavlakosZDD17}&  \checkmark   & {\bf 65.7} &{\bf 70.7}&{\bf  61.6}& 56.4 & {\bf 69.0}& {\bf 59.5} & &{\bf 66.9} & -\\
\hline
Ours (real+mosaic)& \checkmark   & 123.8 & 160.3 & 133.0 & 77.4 & 129.5 & 91.3 &  & 121.2 & 119.5 \\ 
Ours (real+mosaic+surreal)& \checkmark   & 107.4 & 155.2 & 130.7 & 71.6 & 114.6 & 83.1 &  & 110.6 & 112.6 \\ 
\hline
\end{tabular}
}
\caption{Comparison with state of the art on Human3.6M using protocol P2 (average 3D distance without pose
  alignment).  We show per-class results.  Im refers to image-based approaches working at the frame level, i.e., that do not leverage temporal information. 
Note that Du~\etal~\cite{DWLHGWKG16} only evaluate on camera~2.}
\label{tab:H36M_sota2}
\vspace{-2mm}
\end{table*}

   \begin{table*} 
\centering
\begin{tabular}{ccccccc}
\hline
    2D   & 3D   & Number of & Human3.6M & Human3.6M   & Human3.6M   & LSP \\ 
     source &   source & training samples& Abs Error (mm)&   Align. Error (mm) &    Error (pix) &   Error (pix)\\ 
\hline  
Human3.6M   & Human3.6M & 190,000 & 130.1&  97.2 &8.8&31.1 \\  
 MPII+LSP   & Human3.6M &190,000 &248.9& 122.1 &17.3&20.7   \\ 
 MPII+LSP     & CMU & 190,000 &  320.0&150.6 &19.7&22.4  \\ 
 MPII+LSP   & CMU &  $2.10^6$   & 216.5 & 138.0 & 11.2 & 13.8  \\ 
\hline  
\end{tabular} 
\caption{\newtext{Pose error on LSP and Human3.6M using different 2D and 3D sources for rendering the mosaic images and considering different numbers of training samples, i.e., 3D poses and corresponding rendered images.}}
\label{tab:LSPpix}
\end{table*}
We then performed quantitative evaluation by training the same classifier on different combinations of the 3 types of data (real, mosaic and surreal). When combining 2 or 3 types of data, we alternate the batches of each data type considering the exact same poses in the different batches. In practice, we consider batches of 256 images and train for 80k iterations, i.e. 110, 55 and 37 epochs for respectively  1, 2 or 3 types of data in the training set. We evaluate on the same subset of frames from subject S11 that was used in the previous experiments (protocol P1). The numerical results are given in Figure~\ref{fig:numres} where we report classification rate, absolute 3D pose error and 3D pose error after alignment.  We can observe that the model trained on real data (green plot) performs significantly worse than the model trained on our synthetic mosaics (blue) both in terms of classification and 3D pose errors. With even less real data available, subject S9 being removed from the training set for this experiment, our data augmentation process proves to be even more useful. We can also clearly see that the model trained on surreal images (red plot) does not generalize well to new images as the humans are visibly different. This domain gap problem disappears when the surreal images are mixed with real images (yellow curve).  These observations are in par with the conclusions from other works. A domain adaptation process~\cite{ChenWLSWTLCC16,XuRVL14}  or a fine-tuning stage~\cite{HuangR17} is often required when training on purely synthetic human data. Recently, de Souza et al.~\cite{DesouzaGCL17} also showed in the context of action recognition that overfitting can be avoided by training their classifier on a combination of real and synthetic data. Such domain adaptation is not required when using our mosaic data. We can observe that the model trained on mosaic (blue) performs similarly to the model trained on surreal+real data (yellow) and that adding the real data to the mosaic images results in a better model (cyan). Finally, we can see that combining surreal and mosaic images results in an even better model (magenta curve) and that adding real data to the mosaics and surreal data marginally improves the performance  (black). This indicates that our data and the surreal data are complimentary and their association allows to significantly improve the performance when little training data is available. The surreal images bring diversity in terms of body shapes and textures while our mosaics add photo-realism. We believe that we generate hard examples where the symmetry of the body in terms of shape, color and clothing has not been imposed. This seems to help learning more discriminative features.
   \begin{figure*}[htb]
   \centering 
\includegraphics[width=\textwidth]{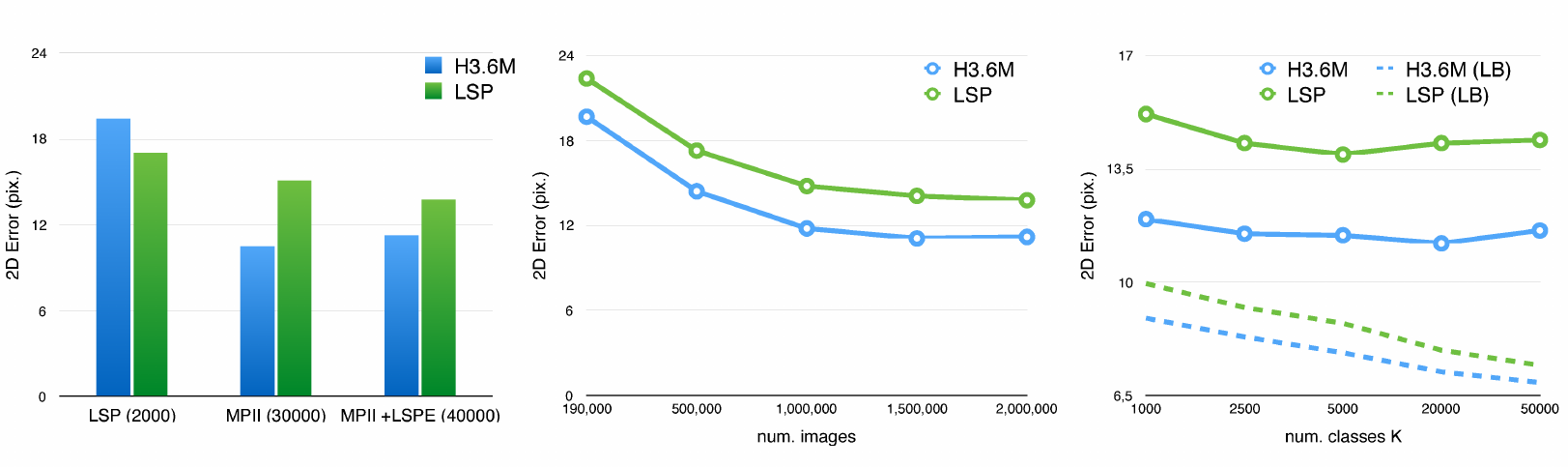}
\caption{ 2D pose error on LSP and
  Human3.6M using different pools of annotated images to generate 2
  millions of mosaic training images (left), varying the number of mosaic training images (center) and considering different number of pose classes $K$ (right). \newtext{On this last plot, we also report the lower-bound (LB) on the 2D error, i.e., computed with the closest classes from ground-truth annotations.}}
     \label{fig:plots}
 \end{figure*} 

In Table~\ref{tab:H36M_sota2}, we compare our results to
  state-of-the-art approaches for the second
  protocol (P2) where
  all the frames from subjects S9 and S11 are used for testing (and
  only S1, S5, S6, S7 and S8 are used for training). Our best
  classifier, trained with a combination of synthetic (mosaic \& surreal) and real data,
  outperforms recent approaches in terms of 3D pose estimation
  for single frames, even methods such as Zhou et al.~\cite{ZhouZLDD16} who integrate temporal information. 
Note that our method estimates absolute pose (including orientation
w.r.t. the camera), which is not the case for other methods such as Bo
et al.~\cite{BoS10}, who estimate a relative pose and do not provide
3D orientation. Only the most recent methods report a better performance~\cite{Moreno17,PavlakosZDD17,RogezWS17,TomeRA17}. They use accurate 2D joint detectors~\cite{Moreno17,TomeRA17} or rely on much more complex architecture~\cite{PavlakosZDD17,RogezWS17} while we employ a simple AlexNet architecture and return a coarse pose estimate.

 \subsection{Evaluation on Leeds Sport Dataset (LSP)}

We now train our pose classifier using different combinations of
training sources and use them to estimate 3D poses on images captured
in-the-wild, i.e., LSP. Since 3D pose evaluation is not possible on
this dataset, we instead compare 2D pose errors expressed in
pixels and measure this error on the normalized $220 \times 220$ images following~\cite{ZhouZLDD16}. We compute the average 2D
pose error over the 13 joints on both LSP and Human3.6M (see
Table~\ref{tab:LSPpix}).   
   \begin{table*} 
\centering
\begin{tabular}{c|cccccccc}
\hline
    Method   & Feet   & Knees & Hips & Hands  & Elbows & Shoulder & Head & All\\ 
\hline  

Wei et al.~\cite{WeiRKS16} &6.6 & 5.3& 4.8& 8.6 & 7.0& 5.2& 5.3& {\bf6.2}\\
Pishchulin et al.~\cite{PishchulinITAAG15} & 10.0 & 6.8 & 5.0 &  11.1 & 8.2 & 5.7 & 5.9 & 7.6\\   
Chen \& Yuille  ~\cite{ChenY14} &15.7 & 11.5& 8.1&  15.6 & 12.1& 8.6  & 6.8 & 11.5\\   
Yang et al.~\cite{yang2016end} &15.5 & 11.5& 8.0 &  14.7 & 12.2& 8.9 & 7.4& 11.5\\
   \hline
Ours (AlexNet) & 19.1 & 13 & 4.9 &  21.4 & 16.6& 10.5 & 10.3& 13.8  \\ 
Ours (VGG)  & 16.2 & 10.6 & 4.1 &  17.7 & 13.0& 8.4 &9.8& 11.5\\  
\hline  
\end{tabular} 
\caption{State-of-the-art results on LSP. The 2D pose error in pixels is computed on the normalized $220 \times 220$ images.}
\label{tab:LSPsota}
\end{table*}

As expected, we observe that when using a pool of the in-the-wild
images to generate the mosaic data, the performance
increases on LSP and drops on Human3.6M, showing the importance of
realistic images for good performance in-the-wild and the lack of
generability of models trained on constrained indoor images. The  error slightly
increases in both cases when using the same number (190,000) of CMU 3D
poses. The same drop was observed by~\cite{IqbalGG16} and can be
explained by the fact that by CMU data covers a larger portions of the
3D pose space, resulting in a worse fit. The results improve on both
test sets when considering more poses and synthetic images (2
millions). The larger drop in Abs 3D error and 2D error compared to  aligned 3D error means that a better 
camera view is estimated when using more synthetic data.
In all cases, the  performance (in pixel) is lower on LSP than on
Human3.6M due to the fact that the poses observed in LSP are more
different from the ones in the CMU motion capture data. In Figure~\ref{fig:plots}
, we visualize the 2D pose error on LSP and Human3.6M 1) for different
pools of annotated 2D images, 2) varying the number of synthesized
training images and 3) considering  different number of pose classes
$K$. As expected using a bigger set of annotated images improves the
performance in-the-wild. Pose error converges both on LSP and Human3.6M
when using 1.5 million of images; using more than $K$=5000 classes does not further improve the performance.
\newtext{The lower-bound on the 2D error, i.e., computed with the closest classes from ground-truth annotations, clearly decreases when augmenting the number $K$ of classes. Smaller clusters and finer pose classes are considered when increasing $K$.  However, the performance does not further increase for larger values of $K$. The classes become probably too similar, resulting in ambiguities in the classification. Another reasons for this observation could be the amount of training data available for each class that also decreases when augmenting $K$.}
  \begin{figure*}[htb]
   \centering 
  \includegraphics[width=0.78\textwidth]{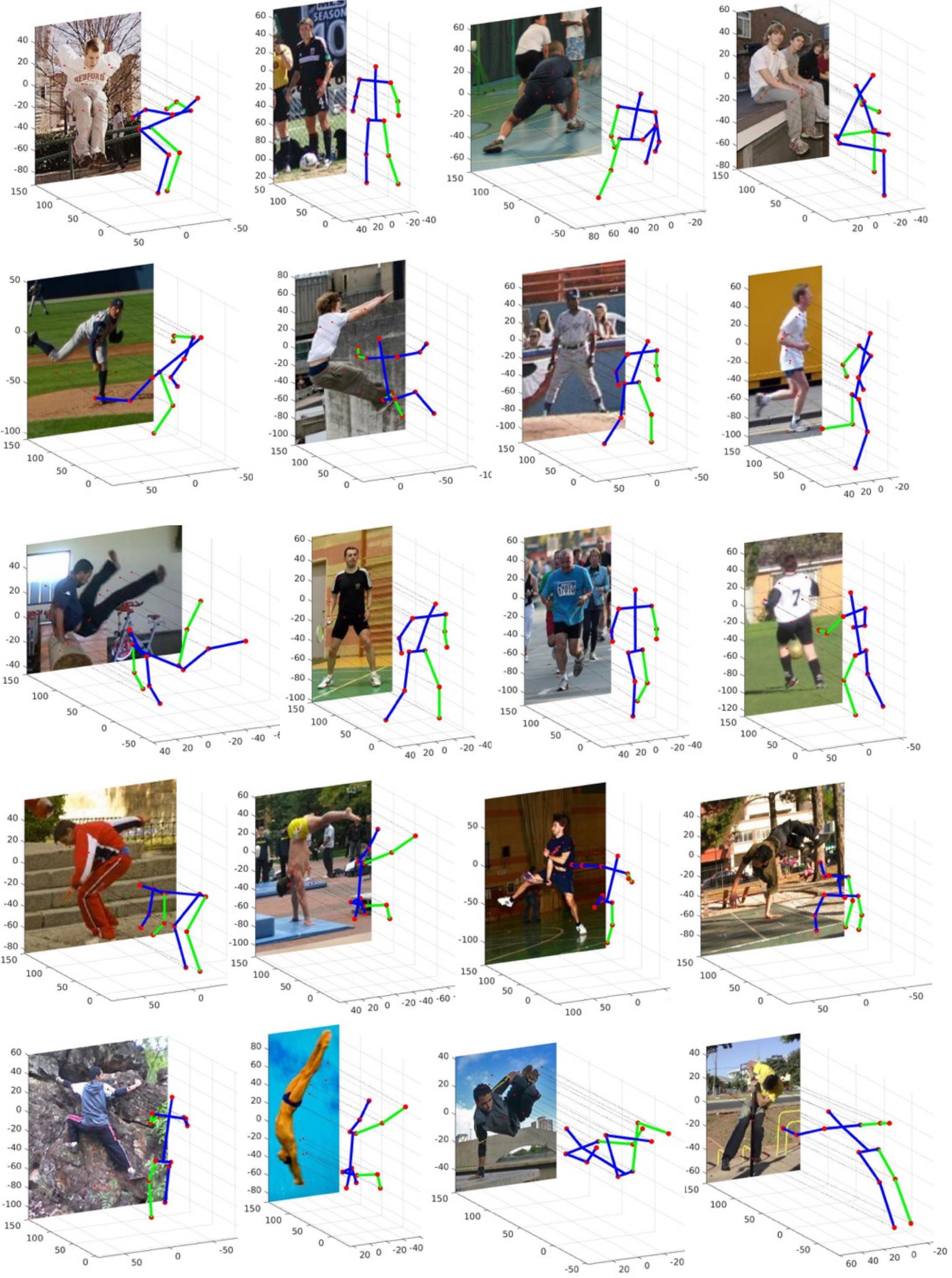}
  \caption{\footnotesize Qualitative results on LSP. We show for each image, the 3D pose corresponding to the top scoring class, i.e. the highest peak in the distribution. 
We show correct 3D pose estimations (top 3 rows), imprecise pose estimation due to coarse discretization (fourth row) and typical failure cases (bottom row), corresponding to unseen poses or right-left and front-back confusions.}
     \label{fig:results}
 \end{figure*}

\noindent {\bf Performance with a deeper architecture. }To further improve the performance, we also experiment with fine-tuning a VGG-16 architecture~\cite{SimonyanZ14a} for pose classification. By doing so, the average (normalized) 2D pose error decreases by $2.3$ pixels.
In Table~\ref{tab:LSPsota}, we compare our results on LSP to the
state-of-the-art 2D pose estimation methods. Although our
approach is designed to estimate a coarse 3D pose, its performances is
comparable to recent 2D pose
estimation methods~\cite{ChenY14,yang2016end}. In Figure~\ref{fig:results}, we present some qualitative results obtained with our method.  For each image, we show the 3D pose corresponding to the average pose of the top scoring class, i.e., the highest peak in the distribution. The qualitative results in Figure~\ref{fig:results} show that our algorithm correctly estimates the global 3D pose.
We also show some failure cases. 
 
\begin{figure*}[htb]
   \centering 
\includegraphics[width=0.8\textwidth]{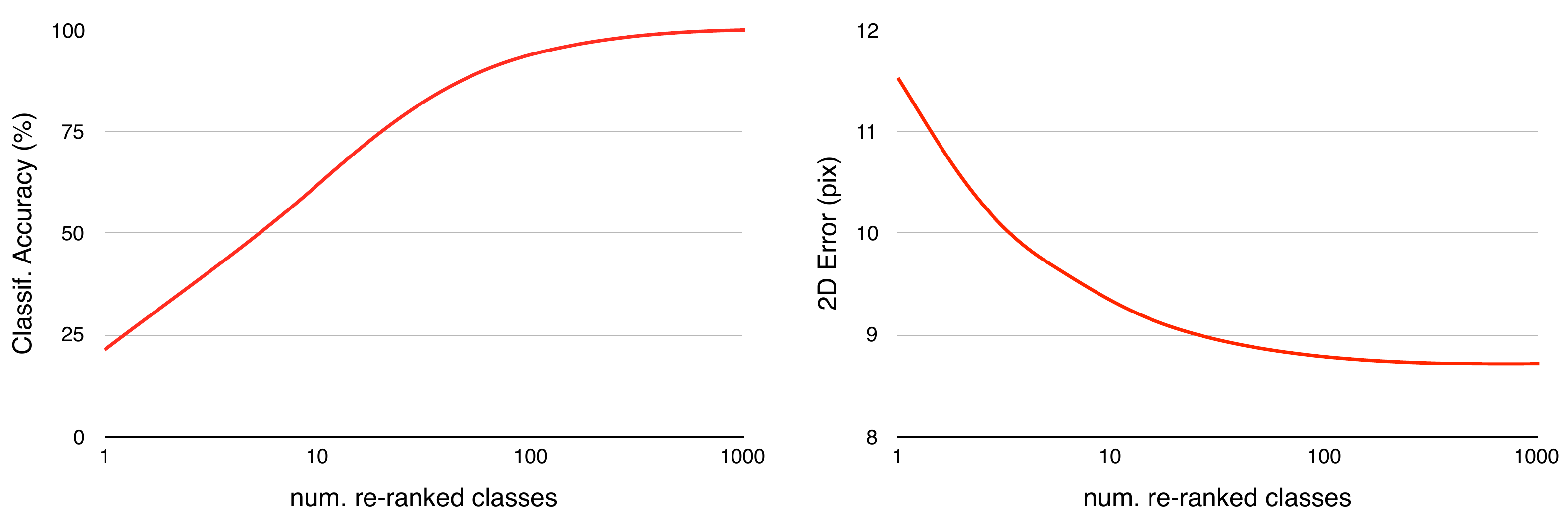}
\caption{ \newtext{Classification accuracy and 2D pose error on LSP when re-ranking the top scoring classes. To show the potential  improvement of a reranking stage, we report the performance when varying the number of top scoring classes re-ranked using ground truth annotations.}}
     \label{fig:plots_reranking}
 \end{figure*}

\noindent {\bf Re-ranking.}  After a visual analysis of the results, we found that failures occur in two cases:  1) when the observed pose does not belong to the motion capture training database, which is a limitation of purely holistic  approaches (e.g., there exists no motion capture 3D pose of a diver as in the second example on the last row in Figure~\ref{fig:results}), or 2) when there is a possible right-left or front-back confusion. We observed that this later case is often correct for subsequent top-scoring poses. \newtext{For the experiments in Table~\ref{tab:LSPsota} using a VGG architecture, the classification rate\footnote{ground truth classes being obtained by assigning the ground truth 2D pose to the closest cluster.} on LSP is  only $21.4\%$, meaning that the classes are very similar and very hard to disambiguate. However, this classification rate reaches $48.5\%$ when considering the best of the 5 top scoring classes, re-ranked using ground truth, as depicted in Fig.~\ref{fig:plots_reranking}. It even reaches $100\%$  when re-ranking the 1000 top scoring classes. i.e., $20\%$ of the $K$=5000 classes. The  2D error  lower bound ($\approx8.7$ pixels) is reached when re-ranking the first 100 top scoring classes, only $2\%$ of the $K$=5000 classes. This highlights a property of our approach that can keep multiple pose hypotheses which could be re-scored adequately, for instance, using temporal information in videos.}

\begin{figure*}[ht]
   \centering 
  \includegraphics[width=\textwidth]{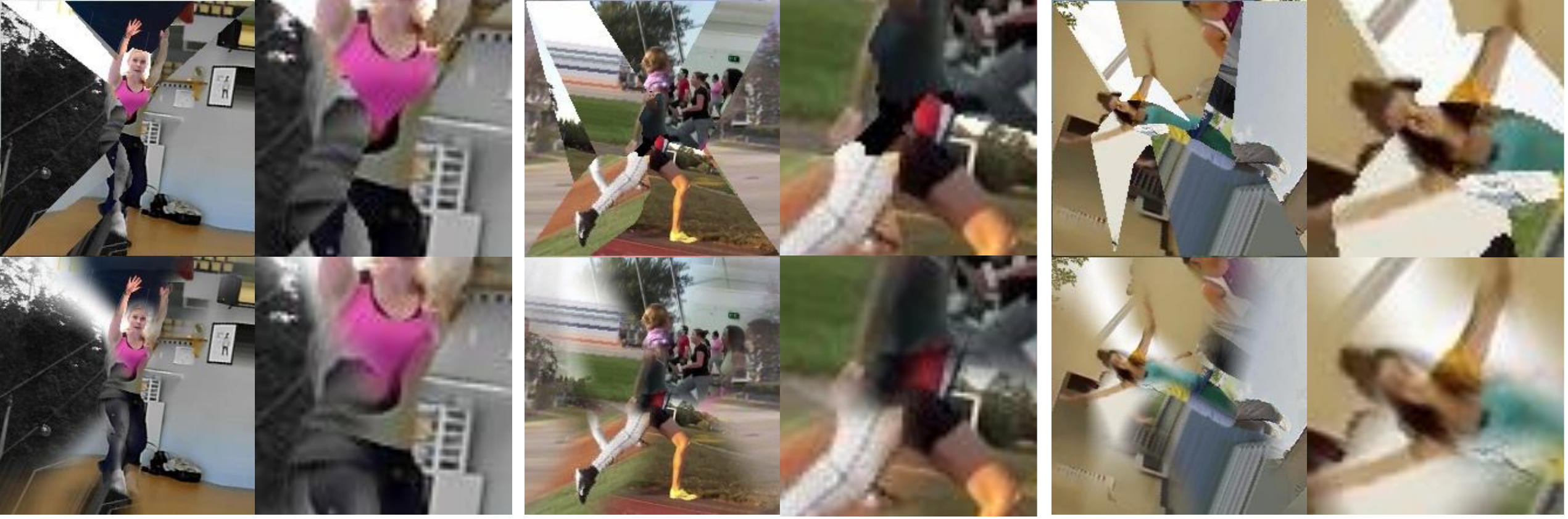}
  \caption{\footnotesize \newtext{Examples of synthetic images generated using  LSP and MPII datasets and CMU motion capture dataset  as 2D and 3D sources respectively. For each case, we show the image before (up) and after (bottom) blending with close-ups.}}
     \label{fig:mosaics_closeups}
 \end{figure*}
\subsection{\newtext{Discussion}}

\newtext{We now analyse the limitations of the proposed method and discuss future research.} 

\begin{figure*}[ht]
   \centering  
     \begin{tabular}{ccc} 
    \multicolumn{3}{c}{\hspace{-2mm}\includegraphics[width=\textwidth]{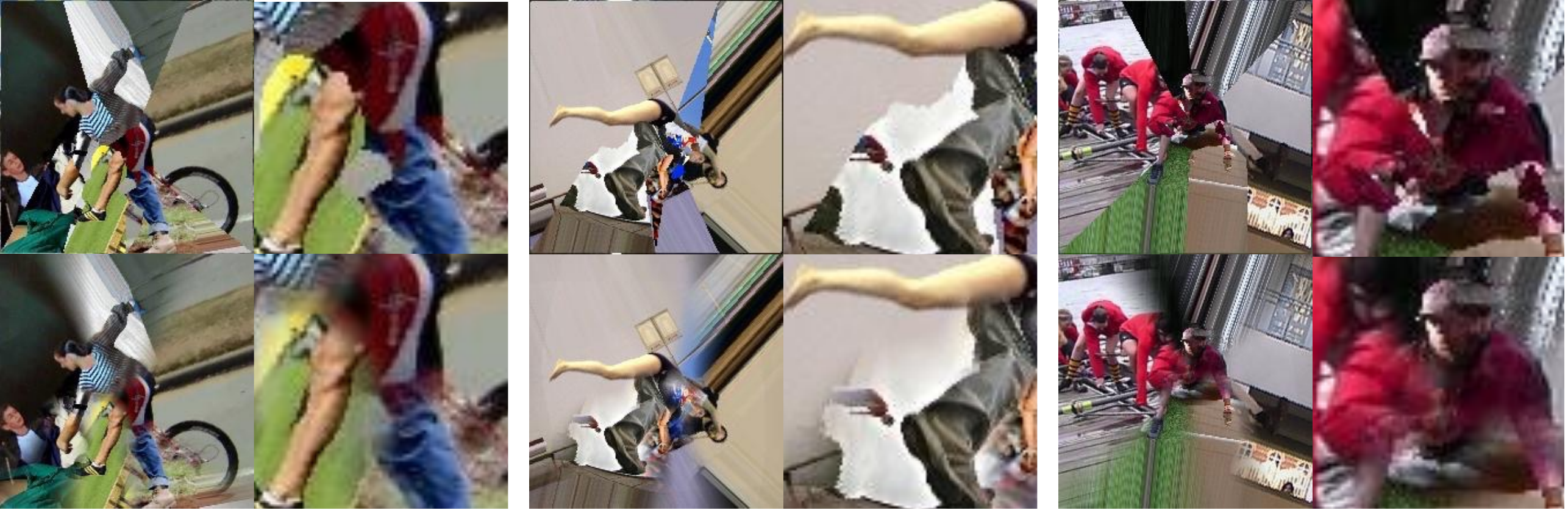}}\\   
 \hspace{24mm} (a)& \hspace{50mm} (b)& \hspace{26mm}(c) 
  \end{tabular}
  \caption{\footnotesize \newtext{Examples of failure cases for synthetic images generated using  LSP and MPII datasets and CMU motion capture dataset  as 2D and 3D sources respectively. For each case, we show the image before (up) and after (bottom) blending with close-ups. We show failures due to differences in clothing (a), 3D depth ambiguity (b) and inaccurate 2D pose annotations (c).}}
     \label{fig:failures}
 \end{figure*}

\noindent \newtext{{\bf Limitations of the method.} In Fig.~\ref{fig:mosaics_closeups}, we show  more visual examples of generated images with our approach before and after blending. To better compare the images with/without blending,  some close-ups are provided. In general, our image-based synthesis engine works well when poses and camera views are similar in the query pose and the annotated images. For instance, if the annotated images only include people observed from the front, our engine will not produce acceptable images from side or top views. In the same way, if the 2D source only contains standing persons, the engine will not be able to synthesise an acceptable image of a sitting pose. If viewpoint and pose are similar in query and annotated images, several factors can influence the quality of the synthesised images. We found three main reasons for failure and show an example of each case in Fig.~\ref{fig:failures}: 1) the similarity in person's morphology and clothing in the selected images, in the example given in Fig.~\ref{fig:failures}a, stitching patches of persons wearing trousers or shorts leads to poor result. 2) the 3D depth ambiguities, this inherent to the fact that matching is performed in 2D and several 3D poses can correspond to the same 2D pose (see Fig.~\ref{fig:failures}b). 3) the quality of the 2D annotations. While the ``perfect'' 2D poses from Human3.6M led to very plausible images, this is not always the case for manual annotations of real images. If the 2D annotations are not consistent or inaccurate, as it often happens with body keypoints such as hips or shoulders,  this can results in a  synthetic image of poor quality  as depicted in Fig.~\ref{fig:failures}c. }

\noindent \newtext{{\bf Pose and views.} A limitation of the proposed approach is that it does not learn the statistics of human poses in real images nor the typical views that can be found in real images. These are two presumably important cues.  Ideally, one would want to synthesise useful training images that better match the test conditions. This could be achieved by sampling from a prior distribution of poses and camera viewpoints instead of randomly selecting them as done in this paper.}

\noindent \newtext{{\bf Image mosaicing.} Our mosaic images do not look realistic. While it seems to be more important for the problem at hand to be locally photorealistic as opposed to globally coherent, this is only true for the 3D pose estimation approaches which were tested in this paper, i.e., deep pipeline with regression or classification objective on images centered on the human. The proposed data might not be appropriate to train more advanced approaches to 3D pose estimation based on a more global reasoning such as~\cite{RogezWS17} who jointly detect the humans in a natural image and estimate their 3D poses. The proposed optimization ignores image compatibility, which it could take into account in future work. If a big enough 2D source, i.e., pool of annotated images, was available, one could constrain the matching in a way that the smallest set of images is used for synthesis, resulting in more satisfactory synthetic images. A new cost function could minimize not just the individual joint scores as a ``unary'' but also some ``binary'' cost that evaluates the match between pairs of joint matches, and also a prior term that encourages color and geometric consistency/minimal images. }
   \begin{table*} 
\centering
\begin{tabular}{ccccc}
\hline
   Pose-aware  & Human3.6M & Human3.6M   & Human3.6M   & LSP \\ 
      Blending    & Abs Error (mm)&   Error (mm) &    Error (pix) &   Error (pix)\\ 
\hline  
yes &  {\bf 320.0}& {\bf 150.6} & {\bf 19.7}& {\bf 22.4  }\\  
no   & 337.6 & 186.2 & 22.8 & 35.9  \\ 
\hline  
\end{tabular} 
\caption{\newtext{Pose error on LSP and Human3.6M with and without blending stage. We compare the results obtained by the AlexNet architecture when trained on  190,000 mosaic images synthesized using CMU motion capture 3D poses and  MPII+LSP dataset as 3D and 2D sources respectively.}}
\label{tab:noBlending}
\end{table*}

\noindent \newtext{{\bf Image blending.} In this work, we have proposed to solve the lack of color and geometric consistency with a pose-aware blending algorithm that removes the artefacts at the boundaries between image regions while maintaining pose informative edges on the person. In Table~\ref{tab:noBlending}, we report the performance of the proposed approach without this  image blending step and show that this second step is actually necessary.
The proposed blending function can seem rather heuristic. Another solution could be a GAN-style~\cite{GoodfellowPMXWOCB14} image synthesis approach: given images and the probability maps, find a generator to generate images that also defeat a discriminative loss. The resulting images would probably look more compelling and probably respect the overall image structure better (coherent body parts and background geometry). This will be explored in future work. Another intriguing question for future research is whether and to what extent a similar approach could generate synthetic videos with 3D annotations. }

 \section{Conclusion}

In this paper, we introduce an approach for creating a synthetic training dataset of  ``in-the-wild'' images and their corresponding 3D pose. Our algorithm artificially augments a dataset of real images with new synthetic
images showing new poses and, importantly, with 3D pose annotations. We showed that CNNs can be trained on these artificially looking images and still  generalize well to real images without requiring any domain adaptation or fine-tuning stage. We train an end-to-end CNN classifier for 3D pose estimation and 
show that, with our synthetic training images, our method outperforms most published methods in terms of 3D pose estimation in controlled environments while employing a much simpler architecture. 
We also demonstrated our approach on the challenging task of estimating  3D body pose of humans in natural images (LSP).  
\newtext{Finally, our experiments highlight that 3D pose classification can outperform regression in the large data regime,  an interesting and not necessarily intuitive conclusion.}
In this paper, we have estimated a coarse 3D pose by returning the
average pose of the top scoring cluster. In future work, we will
investigate how top scoring classes could be re-ranked and also how
the pose could be refined.

\paragraph{\bf Acknowledgments.} 
This work was supported by the European Commission under FP7 Marie Curie IOF grant (PIOF-GA-2012-328288) and partially supported by the ERC advanced  grant ALLEGRO and an Amazon Academic Research Award (AARA).
We acknowledge the support of NVIDIA with the donation of the GPUs
used for this research.
We thank Dr Philippe Weinzaepfel for his help. We also thank the anonymous reviewers for their comments and suggestions that helped improve the paper.

 \footnotesize
\bibliographystyle{plain}
\bibliography{greg_bib,biblio}

\end{document}